%% file: main.tex
\documentclass[10pt,twocolumn,letterpaper]{article}

\usepackage[pagenumbers]{iccv} 

\definecolor{iccvblue}{rgb}{0.21,0.49,0.74}
\usepackage[pagebackref,breaklinks,colorlinks,allcolors=iccvblue]{hyperref}

\title{GeoCD: A Differential Local Approximation for Geodesic Chamfer Distance}

\author{Pedro Alonso$^{1,2}$ \quad Tianrui Li$^{1,2}$ \quad Chongshou Li$^{1,2}$\thanks{Corresponding Author} \\
$^1$ School of Computing and Artificial Intelligence, Southwest Jiaotong University\\
$^2$ Engineering Research Center of Sustainable Urban Intelligent Transportation \\
Ministry of Education, Chengdu, China\\
\texttt{\{palonso, trli, lics\}@swjtu.edu.cn}\\
}

\begin{document}
\maketitle
\input{sec/0_abstract}    
\input{sec/1_intro}
\input{sec/2_related_work}
\input{sec/3_method}
\input{sec/4_experiments}
\input{sec/5_conclusion}
{
    \small
    \bibliographystyle{ieeenat_fullname}
    \bibliography{main}
}

\end{document}

%% file: sec/0_abstract.tex
\begin{abstract}
Chamfer Distance (CD) is a widely adopted metric in 3D point cloud learning due to its simplicity and efficiency. However, it suffers from a fundamental limitation: it relies solely on Euclidean distances, which often fail to capture the intrinsic geometry of 3D shapes. To address this limitation, we propose GeoCD, a topology-aware and fully differentiable approximation of geodesic distance designed to serve as a metric for 3D point cloud learning. Our experiments show that GeoCD consistently improves reconstruction quality over standard CD across various architectures and datasets. We demonstrate this by fine-tuning several models—initially trained with standard CD—using GeoCD. Remarkably, fine-tuning for a single epoch with GeoCD yields significant gains across multiple evaluation metrics.
\end{abstract}

%% file: sec/1_intro.tex
\section{Introduction}
\label{sec:intro}

Point cloud reconstruction is fundamental to tasks such as shape completion \cite{10204938, Liu2019MorphingAS, 9525242, Xiang_2021_ICCV, Yu_2021_ICCV, 10.1109/TPAMI.2023.3309253, yuan2018pcn}, 3D generation \cite{8099747, 10.1007/978-3-030-01237-3_49, 2018arXiv181005795L, 9010395}, and object classification and segmentation \cite{Qi_2017_CVPR, NIPS2017_d8bf84be, 10.1145/3326362}, with applications spanning robotics \cite{9010919}, augmented reality \cite{10.1145/2047196.2047270, 7298631, 10810480, 6162880, 8762161}, and autonomous driving \cite{2018arXiv181205784L, s18103337, 2019arXiv191204838S}. However, due to the unordered and often sparse nature of point clouds, comparing reconstructed and ground truth shapes is a non-trivial. To address this, metrics like Chamfer Distance (CD) and Earth Mover's Distance (EMD) are commonly used.

Chamfer Distance measures the average closest-point distance between two point clouds, while Earth Mover's Distance computes the minimal cost of transforming one point cloud into another. While both metrics have their advantages—CD offers computational efficiency, and EMD enforces one-to-one point correspondences—they share a fundamental limitation: both rely solely on Euclidean distances between point pairs. This reliance limits their ability to capture the intrinsic geometry of 3D shapes, which often lie on curved manifolds that are not well represented in Euclidean space.

To address this limitation, we introduce GeoCD, a differentiable, topology-aware approximation of geodesic distance between points. A geodesic is a generalization of a straight line to curved spaces, defined as the shortest path between two points that lies entirely on the surface of the shape. GeoCD approximates geodesic distances through a multi-hop k-nearest neighbor (kNN) graph. Unlike traditional geodesic methods such as Dijkstra's or Fast Marching algorithms—which are non-differentiable and computationally expensive—GeoCD is fully differentiable and efficient, making it well-suited for integration into deep learning pipelines.

We demonstrate the effectiveness of GeoCD for point cloud reconstruction across multiple architectures and datasets. In our setup, models are first trained using standard CD and then fine-tuned with GeoCD. This simple fine-tuning step leads to consistent improvements in reconstruction quality across several evaluation metrics.

Our contributions are summarized as follows:

\textbullet~ We introduce GeoCD, a differentiable geodesic distance approximation for point cloud learning based on multi-hop kNN graphs.

\textbullet~ We evaluate GeoCD across multiple models and datasets, demonstrating consistent reconstruction improvements.

\textbullet~ We show that a single epoch of GeoCD fine-tuning yields significant gains, providing an effective trade-off between reconstruction quality and computational cost.

%% file: sec/2_related_work.tex
\section{Related Work}
\label{sec:related_work}
\subsection{Point-based Deep Learning}
\label{sec:related_work_dl_for_pc}
Point clouds are inherently unordered and sparse, making them challenging to process with conventional deep learning architectures designed for structured data. Point-based deep learning models address this by operating directly on raw point cloud inputs, avoiding voxelization or multi-view projections, and thereby preserving geometric fidelity and computational efficiency.

A pioneering approach in this direction is PointNet \citep{Qi_2017_CVPR}, which processes each point independently using shared multi-layer perceptrons (MLPs) and aggregates global features through a max-pooling operation. This design achieves invariance to point permutations and enables learning global shape representations directly from unordered point sets. However, PointNet has a key limitation: it fails to capture local geometric relationships because it treats each point independently prior to aggregation. To overcome this, PointNet++ \citep{NIPS2017_d8bf84be} introduces a hierarchical feature learning framework that recursively applies PointNet to local regions and at multiple scales. This hierarchical structure enables the model to capture both local and global shape information more effectively.

DGCNN \citep{10.1145/3326362} further advances local feature learning by introducing EdgeConv, a dynamic graph convolution operator that constructs neighborhoods based on feature-space similarity rather than fixed Euclidean proximity. The graph is recomputed at each layer, allowing the model to adaptively learn context-aware local structures as feature representations evolve. 

More recently, transformer-based architectures have been proposed for point cloud processing. For example, Point Transformer (PT: \cite{zhao2021point}) applies self-attention mechanisms to model contextual relationships among points, enabling each point to dynamically attend to its neighbors. This leads to richer and more flexible representations. Subsequent extensions such as Point Transformer v2 (PTv2: \cite{wu2022point}) and v3 (PTv3: \cite{wu2024point}) further improved the attention mechanism, scalability, and positional encoding strategies. In particular, PTv3 introduced architectural refinements that significantly enhance computational efficiency and achieve state-of-the-art performance on various 3D vision benchmarks.

\subsection{Chamfer Distance}
\label{sec:related_work_chamfer_distance}
Chamfer Distance (CD) is one of the most widely used metrics for comparing point clouds. It measures the average squared distance between each point in one set and its nearest neighbor in the other. Given two point clouds P and Q, CD is defined as,

\begin{equation}
    \text{CD}(P, Q) = \frac{1}{N_P} \sum_{p \in P} \min_{q \in Q} \|p - q\|^2 + \frac{1}{N_Q} \sum_{q \in Q} \min_{p \in P} \|q - p\|^2,
    \label{eq:cd}
\end{equation}

where $N_P$ and $N_Q$ are the number of points in P and Q, respectively.

CD is popular due to its simplicity, differentiability, and computational efficiency, making it well-suited for training deep learning models. However, it also has several well-known limitations: (1) it is sensitive to outliers, (2) it does not enforce one-to-one point correspondence, which can lead to density imbalances, and (3) it relies solely on Euclidean distances, failing to capture the intrinsic geometry of curved surfaces.

To address these issues, several extensions of CD have been proposed. For example, Density-aware Chamfer Distance (DCD: \cite{10.5555/3540261.3542489}) introduces point-wise weights based on local density, penalizing redundancy while emphasizing undersampled areas.

More recently, \citep{Lin2023HyperbolicCD} proposed the Hyperbolic Chamfer Distance (HyperCD), which extends the standard Chamfer Distance to hyperbolic space. HyperCD incorporates a position-aware weighting mechanism derived from hyperbolic geometry, enhancing the modeling of local structures and improving robustness to outliers, resulting in smoother and more accurate reconstructions.

%% file: sec/3_method.tex
\section{Method}
\label{sec:method}

\begin{figure*}
  \centering
  \includegraphics[width=1\textwidth]{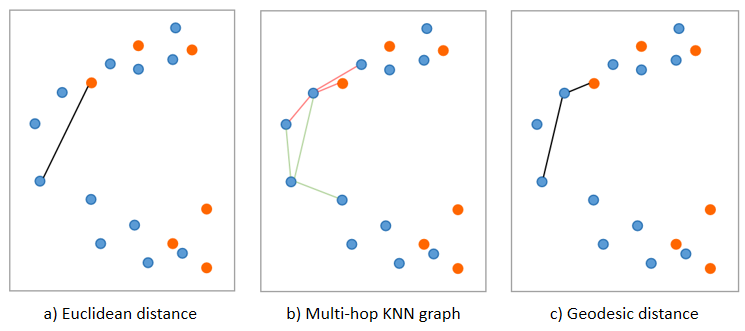}
  \caption{Illustration of the geodesic distance approximation. (a) Euclidean distance between two points. (b) kNN graph for $k=3$. (c) Multi-hop geodesic approximation, where the same two points are now connected via a 2-hop path through an intermediate node.}
  \label{fig:geocd_hops}
\end{figure*}

To approximate geodesic distances on unstructured point clouds in a differentiable way, we construct a multi-hop $k$-nearest neighbors (kNN) graph over the combined set of predicted and ground truth points. This graph is then used to propagate local distances across multiple hops, resulting in a more topology-aware distance metric.
\subsection{Multi-hop kNN Graph Construction}
\label{sec:multi-hop}

Given a predicted point cloud $X \in \mathbb{R}^{B \times N \times 3}$ and ground truth point cloud $Y \in \mathbb{R}^{B \times M \times 3}$, we concatenate them into a unified set $Z = X \cup Y \in \mathbb{R}^{B \times (N+M) \times 3}$. For each point $z_i \in Z$, we compute its $k$ nearest neighbors in Euclidean space to construct a local connectivity graph. This defines the 1-hop graph $\mathcal{G}^{(1)}$, whose edges are weighted by the Euclidean distances between neighboring points.

Formally, let $A^{(1)} \in \mathbb{R}^{(N+M) \times (N+M)}$ be the adjacency matrix of $\mathcal{G}^{(1)}$:
\begin{equation}
A^{(1)}[i, j] = 
\begin{cases}
\|z_i - z_j\|_2 & \text{if } z_j \in \text{kNN}(z_i), \\
\infty & \text{otherwise}.
\end{cases}
\label{eq:matrix_1}
\end{equation}

In practice, for numerical stability, we set non-neighbor entries to 1 instead of $\infty$. Because the point clouds are normalized to a unit bounding box in our experiments, all true pairwise distances are smaller than 1. This ensures that non-neighbor distances contribute minimally to the softmin operation (see Section~\ref{sec:softmin}) while avoiding numerical instability.

The multi-hop propagation proceeds iteratively: the 2-hop distances $A^{(2)}$ are computed by finding the shortest paths of length two, and so on. At each iteration $h$, we compute a new matrix $A^{(h)}$ that captures the shortest known distances between all point pairs using paths of up to $h$ hops. The update rule is given by:

\begin{align}
A^{(h)}[i, j] 
&= \min \left[ A^{(h-1)}[i, j],\right. \notag \\
&\quad\left. \min_k \left( A^{(h-1)}[i, k] + A^{(1)}[k, j] \right) \right].
\label{eq:matrix_2}
\end{align}

This update is a form of min-plus matrix multiplication, where standard addition and multiplication operations are replaced by addition and minimum, respectively. Unlike traditional algorithms such as Dijkstra's, which operate sequentially from a single source node, our min-plus updates compute the shortest paths between all pairs of points in a parallel and differentiable manner, enabling efficient batch processing during training.

By iterating this process up to $n$ hops, we allow information to propagate through increasingly longer paths while preserving locality. Each additional hop incorporates more distant relationships within the point cloud, enabling GeoCD to better align points that are close in geodesic space but far apart in Euclidean space.

Figure~\ref{fig:geocd_hops} illustrates the key intuition behind our geodesic distance approximation. Blue and orange points represent two separate point clouds after merging. (a) shows the standard Euclidean distance between two points, as used in traditional Chamfer Distance. (b) depicts the multi-hop kNN graph construction with k=3. Finally, (c) highlights the geodesic path approximated via the multi-hop graph: the same two points are now connected through an intermediate node (a 2-hop path), which more closely follows the underlying surface geometry compared to the straight-line Euclidean distance.

Finally, after constructing the complete distance matrix $A^{(n)}$ over all points in $Z$, we extract the $N \times M$ submatrix corresponding to distances between predicted points in $X$ and ground truth points in $Y$. These cross-set distances are then used as input to our softmin operation (see Equation~\ref{eq:geocd_loss}).

\subsection{Softmin Operation}
\label{sec:softmin}
Unlike the standard Chamfer Distance, where the nearest-neighbor operation remains differentiable almost everywhere, our geodesic approximation requires a smoother alternative. Because distances are propagated through multi-hop graph updates, many entries are equal or close to 1 for non-neighbors, and the hard minimum introduces discontinuities that destabilize optimization.

To ensure smooth gradients and maintain differentiability, we replace the hard nearest-neighbor matching in Chamfer Distance with a \textit{softmin} operation. For a point $x_i$ in the predicted point cloud, its softmin distance to the ground truth is defined as,

\begin{equation}
\text{softmin}(d_i) = -\log \sum_j \exp(-d_{ij}),
\end{equation}
where $d_{ij}$ denotes the distance between point $x_i$ and point $y_j$. This formulation provides a smooth approximation to the minimum, ensuring that gradients can be computed reliably during training.

\subsection{GeoCD Loss}
\label{sec:GeoCD}

Let $D_{\text{geo}} \in \mathbb{R}^{B \times (N+M) \times (N+M)}$ be the final geodesic distance matrix computed from the $n$-hop graph. The GeoCD loss is then defined as,

\begin{equation}
\label{eq:geocd_loss}
\begin{split}
\mathcal{L}_{\text{GeoCD}}(X, Y) = & \frac{1}{N} \sum_{i=1}^{N} \text{softmin}(D_{X \rightarrow Y}[i, :]) \\
+\, & \frac{1}{M} \sum_{j=1}^{M} \text{softmin}(D_{Y \rightarrow X}[j, :]).
\end{split}
\end{equation}

Here, $D_{X \rightarrow Y}$ denotes the submatrix of $D_{\text{geo}}$ containing distances from predicted to ground truth points.

%% file: sec/4_experiments.tex
\section{Experiments}
\label{sec:experiments}

\begin{table*}
  \caption{Quantitative evaluation of point cloud reconstruction on the ModelNet40 and ShapeNetPart datasets using two architectures: a simple Autoencoder (AE) and Point Transformer v3 (PTv3). We report results for models trained with Chamfer Distance (CD) as well as the same models fine-tuned with GeoCD. Evaluation metrics include Chamfer Distance (CD), Hausdorff Distance (HD), and F1-score at 1\% threshold (F1@1\%). Changes in performance for fine-tuned models (highlighted in green) are relative to the same model without fine-tuning.}
  \label{tab:results_table}
  \centering
  \begin{tabular}{lccc}
    \toprule
    \multicolumn{4}{c}{\textbf{ModelNet40}}\\
    \midrule
    \textbf{Model} & \textbf{$10^3\times$CD} 
                      & \textbf{$10^2\times$HD} 
                      & \textbf{F1@1\%}\\
    \midrule
    AE (no fine-tuning) & 3.42  & 16.25 & 26.48 \\
    AE (GeoCD fine-tuned) & \textbf{3.32} (\textcolor{green}{$\uparrow$0.10}) & \textbf{15.85} (\textcolor{green}{$\downarrow$0.40}) & \textbf{28.24} (\textcolor{green}{$\uparrow$1.76}) \\
    \midrule
    PTv3 (no fine-tuning) & 2.85 & 15.61 & 29.01 \\
    PTv3 (GeoCD fine-tuned) & \textbf{2.77} (\textcolor{green}{$\uparrow$0.08}) & \textbf{15.55} (\textcolor{green}{$\downarrow$0.06}) & \textbf{30.92} (\textcolor{green}{$\uparrow$1.91}) \\
    \midrule
    \multicolumn{4}{c}{\textbf{ShapeNetPart}}\\
    \midrule
    AE (no fine-tuning) & 2.52 & 14.25 & 38.38 \\
    AE (GeoCD fine-tuned) & \textbf{2.46} (\textcolor{green}{$\uparrow$0.06}) & \textbf{13.69} (\textcolor{green}{$\downarrow$0.56}) & \textbf{40.14} (\textcolor{green}{$\uparrow$1.06}) \\
    \midrule
    PTv3 (no fine-tuning) & 2.24 & 13.33 & 40.83 \\
    PTv3 (GeoCD fine-tuned) & \textbf{2.20} (\textcolor{green}{$\uparrow$0.04}) & \textbf{13.16} (\textcolor{green}{$\downarrow$0.17}) & \textbf{42.26} (\textcolor{green}{$\uparrow$1.43}) \\
    \bottomrule
  \end{tabular}
\end{table*}

\subsection{Experimental Setup}
\subsubsection{Datasets}
We evaluate our models on two widely used benchmarks for point cloud analysis: ModelNet40 and ShapeNetPart.

\textbf{ModelNet40: }ModelNet40 \citep{Wu_2015_CVPR} contains 12,311 synthetic 3D shapes across 40 categories, with 9,840 samples for training ($80\%$) and 2,468 for testing ($20\%$). The dataset provides clean, uniformly sampled, and aligned point clouds, making it a standard benchmark for tasks such as 3D shape classification and reconstruction.

\textbf{ShapeNetPart: }ShapeNetPart \citep{10.1145/2980179.2980238} consists of a total of 12,137 shapes for training ($70\%$), 1,870 for validation ($10\%$), and 2,874 for testing ($20\%$), divided into 16 different categories. It is derived from the larger ShapeNet dataset \citep{chang2015shapenetinformationrich3dmodel} and provides part-level annotations, making it especially suitable for part segmentation tasks.

\subsubsection{Models}

\begin{figure*}
  \centering
  \includegraphics[width=0.58\textwidth]{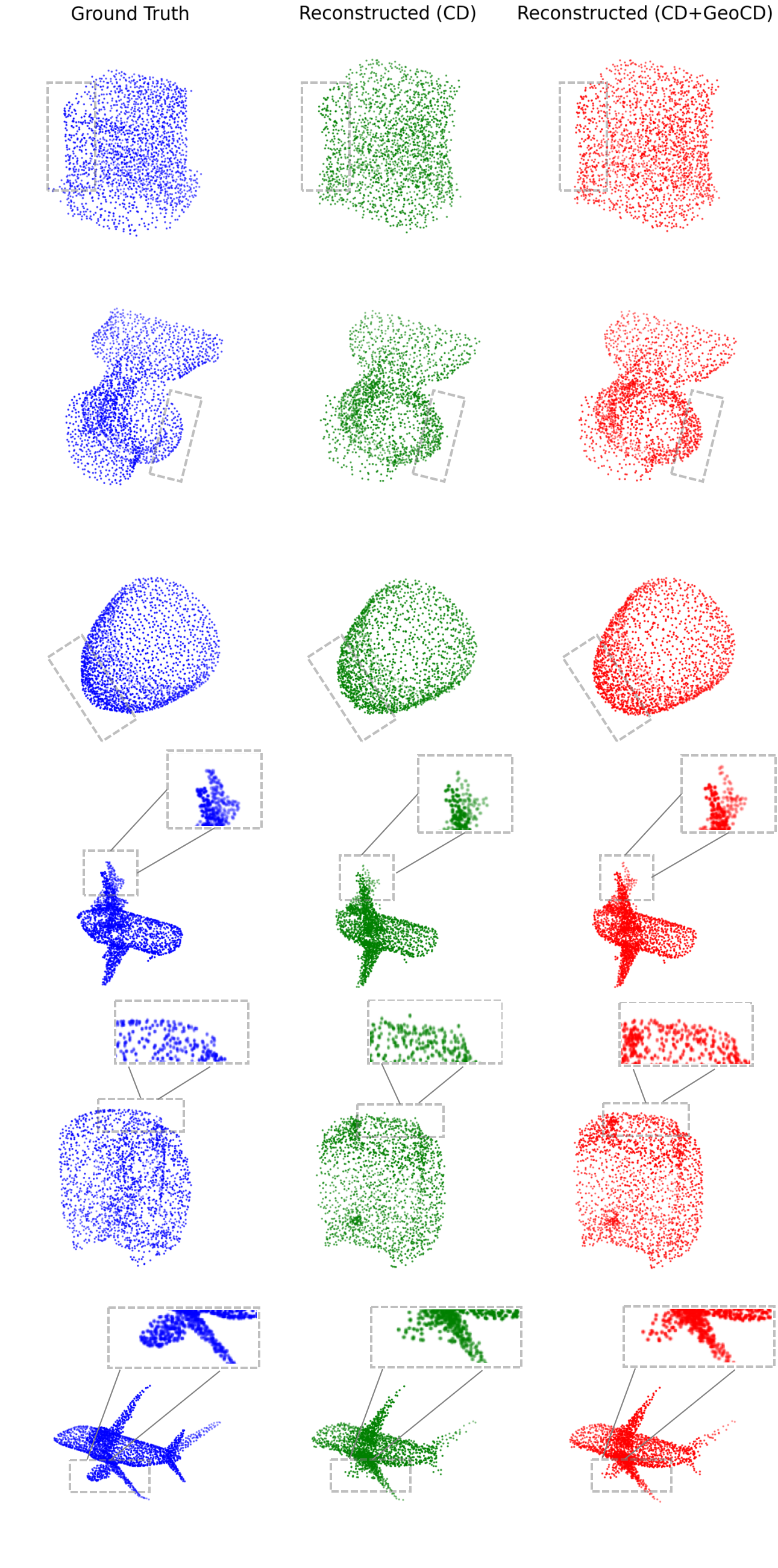}
  \caption{Qualitative comparison of point cloud reconstructions. Each row shows a ground truth shape (left), the output of a model trained only with CD (middle), and the output of the same model fine-tuned with GeoCD (right). Regions with noticeable improvements are highlighted with grey boxes.}
  \label{fig:visual_results}
\end{figure*}

We perform experiments across two backbone architectures: a simple autoencoder (AE) and Point Transformer v3 (PTv3).

\textbf{Autoencoder: } The autoencoder consists of three encoder layers implemented as 1D convolutional layers with channel dimensions $3\to64$, $64\to128$, and $128\to128$. Each layer is followed by batch normalization and ReLU activation, except for the final encoder layer, which omits ReLU. From the resulting latent vector of dimension 128, three decoder layers reconstruct the point cloud back to its original size, with channel dimensions $ 128 \to 256 \to 512 \to N\times3$, where $N$ is the number of points in the point cloud (2048 in our experiments).

\textbf{PTv3: } Point Transformer v3 (PTv3; \citep{wu2024point}) is a state-of-the-art transformer-based architecture for point cloud processing. It builds upon the foundations laid by Point Transformer v1 (PTv1: \cite{zhao2021point}) and v2 (PTv2: \cite{wu2022point}). PTv1 introduced a self-attention mechanism designed for point clouds, enabling the model to capture rich local geometric relationships. PTv2 extended this approach by incorporating rotation-invariant relative positional encoding. PTv3 further improves the architecture by enhancing scalability and training stability, achieving state-of-the-art performance across various 3D vision tasks.

\subsubsection{Training Details}

Models are trained for 100 epochs using Chamfer Distance (eq.~\ref{eq:cd}) as the objective. We then select the best-performing model according to evaluation loss and fine-tune it for a single epoch using GeoCD (see Section~\ref{sec:ablation_nepochs}). We use the Adam optimizer \citep{kingma2014adam}. The learning rate is set to $5\cdot10^{-4}$ for AE and $1\cdot10^{-4}$ for PTv3, and remains unchanged during fine-tuning. To ensure fair comparisons, we use a batch size of 8 for all experiments, as GeoCD significantly increases memory consumption. For the kNN graph, we use $k=5$ (see Section~\ref{sec:ablation_k}), and we calculate up to 2 hops.

All experiments were conducted on a single NVIDIA RTX 4090 GPU.

In practice, Equation~\ref{eq:matrix_2} can be computationally expensive, as it involves operations on distance matrices of size $B\times (N+M)\times (N+M)$, where $B$ is the batch size and $N$ and $M$ are the number of points in the predicted and ground truth point clouds, respectively. For point clouds containing thousands of points, computing full min-plus matrix products over all pairs becomes infeasible due to memory and runtime constraints.

To address this, we introduce a masking strategy during multi-hop propagation. Specifically, once a point is within a certain distance threshold from any point in the other point cloud, it is masked out from further updates in subsequent hops. This approximation does not affect the theoretical formulation, since any longer path involving additional hops would only increase the distance and therefore would not contribute to the minimum in Equation~\ref{eq:matrix_2}.  

\subsubsection{Evaluation Metrics}

We evaluate point cloud reconstruction quality using the Chamfer Distance (see Section~\ref{sec:related_work_chamfer_distance}) and two additional standard metrics commonly employed in point cloud analysis:

\textbf{Hausdorff Distance (HD): } Measures the largest minimal distance between two point sets. Given point sets P and Q, the Hausdorff Distance is defined as,

\begin{equation}
    \text{HD}(P, Q) = \max \left\{ \sup_{p \in P} \inf_{q \in Q} \|p - q\|_2,\; \sup_{q \in Q} \inf_{p \in P} \|q - p\|_2 \right\},
\end{equation}
where $\|\cdot\|$ denotes the Euclidean norm.

\textbf{F1 Score: } Measures the overlap between predicted and ground truth points by combining precision and recall. It is defined as,

\begin{equation}
    \text{F1} = \frac{2 \cdot \text{Precision} \cdot \text{Recall}}{\text{Precision} + \text{Recall}}.
\end{equation}

In this work, we report the F1-score at a $1\%$ threshold ($F1@1\%$). A point is considered a true positive (TP) if it lies within 1\% of the bounding box diagonal from its nearest neighbor in the other point cloud. Otherwise, it is counted as a false positive (FP).

\subsection{Main Results}

Table~\ref{tab:results_table} summarizes our point cloud reconstruction results across two models (AE and PTv3) and two datasets (ModelNet40 and ShapeNetPart). Performance is evaluated using Chamfer Distance (CD), Hausdorff Distance (HD), and F1 score at a 1\% threshold ($F1@1\%$). We observe consistent improvements across all metrics when fine-tuning with GeoCD compared to models trained only with CD. All models were fine-tuned for a single epoch, as additional training yielded negligible gains (see Section~\ref{sec:ablation_nepochs}).

Qualitative reconstruction results are presented in Figure~\ref{fig:visual_results}. In each row, the ground truth point cloud is shown on the left, followed by the output of the model trained only with CD and the output of the same model fine-tuned with GeoCD. Regions with noticeable improvements are highlighted with grey boxes. Although visual differences are sometimes subtle, fine-tuned models generally produce smoother surfaces that follow the underlying geometry more accurately, especially around curved regions and object boundaries.

\begin{table*}
  \caption{Impact of $k$ on reconstruction metrics and runtime for the autoencoder evaluated on ModelNet40. We report Chamfer Distance (CD), Hausdorff Distance (HD), and F1 score. Time per batch refers to the average training time per iteration.}
  \label{tab:k_comparison_table}
  \centering
  \begin{tabular}{lcccc}
    \toprule
    \multicolumn{5}{c}{\textbf{ModelNet40}}\\
    \midrule
    \textbf{Model} & \textbf{$10^3\times$CD} 
                      & \textbf{$10^2\times$HD} 
                      & \textbf{F1@1\%}
                      & \textbf{Time per batch (s)}\\
    \midrule
    AE (k=3) & 3.31  & 15.85 & 28.47 & 20.48 \\
    AE (k=5) & 3.32  & 15.85 & 28.24 & 19.69 \\
    AE (k=10) & 3.34  & 15.90 & 28.04 & 20.78 \\
    \bottomrule
  \end{tabular}
\end{table*}

\subsection{Ablation Studies}
\subsubsection{Impact of $k$ on Graph Connectivity and Performance}
\label{sec:ablation_k}
The parameter $k$ plays a crucial role in GeoCD, as it determines the number of neighboring points used to construct the kNN graph. A small $k$ can limit connectivity, increasing the risk that some points remain disconnected or require many hops to establish a geodesic path, which significantly increases computational cost. In contrast, a large $k$ increases the likelihood that the nearest neighbors include points from the other point cloud, causing many distances to be resolved in the first hop and effectively reverting to Euclidean distance.

We evaluated the impact of $k$ by training the autoencoder model on ModelNet40 with three different values: $k=3$, $k=5$, and $k=10$. Table~\ref{tab:k_comparison_table} reports reconstruction performance across Chamfer Distance (CD), Hausdorff Distance (HD), and F1-score, along with the average training time per batch.

Results show that $k=3$ and $k=5$ achieve comparable reconstruction accuracy, while $k=10$ yields slightly worse performance. In terms of runtime, $k=5$ provides a modest reduction in computational time compared to the other configurations. To balance the risk of missing connections and the need for efficient training, we adopt $k=5$ as the default setting in all main experiments.

\begin{table*}
  \caption{Effect of fine-tuning epochs on reconstruction metrics. Results are reported for the autoencoder trained on ModelNet40.}
  \label{tab:nepochs_table}
  \centering
  \begin{tabular}{lccc}
    \toprule
    \multicolumn{4}{c}{\textbf{ModelNet40}}\\
    \midrule
    \textbf{Model} & \textbf{$10^3\times$CD} 
                      & \textbf{$10^2\times$HD} 
                      & \textbf{F1@1\%} \\
    \midrule
    AE (1 epoch fine-tuning) & 3.32  & 15.85 & 28.24 \\
    AE (2 epochs fine-tuning) & 3.32  & 15.87 & 28.27 \\
    AE (3 epochs fine-tuning) & 3.32  & 15.83 & 28.43 \\
    \bottomrule
  \end{tabular}
\end{table*}

\subsubsection{Effect of Fine-tuning Duration}
\label{sec:ablation_nepochs}
We assess the impact of fine-tuning duration by evaluating the autoencoder on ModelNet40 after different numbers of fine-tuning epochs with GeoCD. The model is initially trained for 100 epochs using the standard Chamfer Distance, and subsequently fine-tuned with GeoCD for 1, 2, and 3 epochs. Results are summarized in Table~\ref{tab:nepochs_table}.

We observe no significant improvements beyond the first epoch of fine-tuning. While the F1-score increases marginally with additional epochs, the Chamfer and Hausdorff distances remain effectively unchanged. Given the substantial computational cost of each GeoCD epoch (see Section~\ref{sec:discussion_2}), fine-tuning for a single epoch offers the most practical balance between efficiency and reconstruction quality.

\subsection{Discussion}
\label{sec:discussions}
\subsubsection{Training Strategy}
\label{sec:discussion_1}
Our multi-hop kNN method relies on a reasonably good alignment between the ground truth and the reconstructed point clouds. This is because both point sets are merged into a single set prior to constructing the graph, and the geodesic distances are defined over this combined surface. If the predictions are poorly aligned or randomly distributed, the resulting graph connections do not correspond to any meaningful geometric paths between the two shapes. For this reason, GeoCD is applied only during the fine-tuning phase of our experiments, after the model has first been trained with standard Chamfer Distance to achieve a coarse alignment.

\subsubsection{Computational Time}
\label{sec:discussion_2}
GeoCD introduces a significant computational overhead compared to the standard Chamfer Distance. For example, training for 100 epochs with CD takes approximately 5 minutes for the autoencoder and around 8 hours for PTv3 on an NVIDIA RTX 4090 GPU. In contrast, fine-tuning for a single epoch with GeoCD takes approximately 6 hours for the autoencoder and 7 hours for PTv3 under the same conditions.

While the improvements from GeoCD fine-tuning are substantial for both models, the additional computational cost per epoch is comparable only for PTv3 to the total training time of CD. Interestingly, GeoCD runtime does not strongly depend on the model complexity, resulting in similar training times for both simple (AE) and complex (PTv3) architectures. This suggests that GeoCD may be especially suitable for more complex models, where its relative overhead is less pronounced compared to baseline training costs.

The main factor contributing to this increased runtime is the cost of constructing the multi-hop kNN graph and performing repeated min-plus updates over the full pairwise distance matrix. Even with practical optimizations such as masking strategies and batch processing, the computational cost remains high.

%% file: sec/5_conclusion.tex
\section{Conclusion}
\label{sec:conclusion}
In this paper, we address a fundamental limitation of the Chamfer Distance: it relies solely on Euclidean distances between points, which fail to capture the intrinsic geometry of curved surfaces. To overcome this, we introduce GeoCD, a fully differentiable method for approximating geodesic distances in point clouds. GeoCD constructs a multi-hop kNN graph to generate paths between points over several hops. Our method is fully differentiable, making it well-suited for integration into deep learning training pipelines. We conduct extensive experiments across multiple models and datasets. Models trained with the standard Chamfer Distance and fine-tuned with GeoCD show consistent improvements across several metrics compared to those trained without fine-tuning.

\textbf{Limitations. }

While GeoCD presents consistent improvements in point cloud reconstruction, it still faces several limitations:

\textbullet~ \textbf{Fine-tuning vs full training: } In the current approach, GeoCD is only used for fine-tuning. As discussed in Section~\ref{sec:discussion_1}, GeoCD requires a reasonably good alignment between the ground truth and reconstructed point clouds, since geodesic distances are computed over the merged set. Extending GeoCD to support effective training from scratch remains an open direction for future work.

\textbullet~ \textbf{Computational overhead: } As described in Section~\ref{sec:discussion_2}, the computational overhead of GeoCD can be substantial, particularly for simple models such as our autoencoder. This challenge is likely to be even more pronounced when applied during full training, thus reducing computational cost will be essential for broader adoption.

\textbullet~ \textbf{Memory cost: } Another limitation is increased memory usage. Although we have introduced practical optimization strategies such as masking, GeoCD fine-tuning still requires more memory than training with the standard Chamfer Distance.

In future work, we plan to explore more efficient approximations of geodesic paths and investigate strategies to integrate GeoCD as the primary loss during full training. Moreover, reducing memory and runtime overhead will be essential for scaling to more complex models and larger point clouds.